\documentclass[10pt,twocolumn,letterpaper]{article}

\usepackage[pagenumbers]{cvpr} 
\usepackage{capt-of}
\usepackage{etoolbox}

\usepackage[dvipsnames]{xcolor}

\definecolor{cvprblue}{rgb}{0.21,0.49,0.74}
\usepackage[pagebackref,breaklinks,colorlinks,citecolor=cvprblue]{hyperref}
\usepackage{pifont}
\newcommand{\xmark}{\ding{55}}
\usepackage{multirow}
\usepackage{diagbox}
\newcommand{\ra}[1]{\renewcommand{\arraystretch}{#1}}

\def\paperID{930} 
\def\confName{CVPR}
\def\confYear{2024}

\title{DETER: Detecting Edited Regions for Deterring Generative Manipulations}

\author{Sai Wang*$^{1}$, Ye Zhu*$^{2}$, Ruoyu Wang$^1$, Amaya Dharmasiri$^2$, Olga Russakovsky$^2$, Yu Wu$^1$\\
$^1$School of Computer Science, Wuhan University \\
$^2$Department of Computer Science, Princeton University\\
{\tt\small wangsai23@whu.edu.cn$^*$, yezhu@princeton.edu$^*$, wangruoyu@whu.edu.cn,} \\ {\tt\small dk9893@princeton.edu, olgarus@princeton.edu, wuyucs@whu.edu.cn}\\
{\small $*$ for equal contributions}\\
\small Project page at \href{https://deter2024.github.io/deter/}{https://deter2024.github.io/deter/}}

\begin{document}
\maketitle

\begin{abstract}
Generative AI capabilities have grown substantially in recent years, raising renewed concerns about potential malicious use of generated data, or ``deep fakes.’’ 
However, deep fake datasets have not kept up with generative AI advancements sufficiently to enable the development of deep fake detection technology which can meaningfully alert human users in real-world settings. Existing datasets typically use GAN-based models and introduce spurious correlations by always editing similar face regions. 
To counteract the shortcomings, we introduce \textbf{DETER}, a large-scale dataset for \textbf{DETE}cting edited image \textbf{R}egions and \textbf{deter}ring modern advanced generative manipulations. \textbf{DETER} includes 300,000 images manipulated by four state-of-the-art generators with three editing operations: face swapping (a standard coarse image manipulation), inpainting (a novel manipulation for deep fake datasets), and attribute editing (a subtle fine-grained manipulation). While face swapping and attribute editing are performed on similar face regions such as eyes and nose, the inpainting operation can be performed on random image regions, removing the spurious correlations of previous datasets. Careful image post-processing is performed to ensure deep fakes in \textbf{DETER} look realistic, and human studies confirm that human deep fake detection rate on DETER is 20.4\% lower than on other fake datasets. Equipped with the dataset, we conduct extensive experiments and break-down analysis using our rich annotations and improved benchmark protocols, revealing future directions and the next set of challenges in developing reliable regional fake detection models.

\end{abstract}    
\vspace{-0.1in}
\section{Introduction}
\label{sec:intro}

\textit{``If you know both enemy and yourself, you will fight a hundred battles without danger or failure."} 

\textit{\hfill -- Sun Tzu, The Art of War, 300 BC}

Generative AI models such as StableDiffusion~\cite{stablediff} and ChatGPT~\cite{OpenAI_ChatGPT} have captured significant attention from both the research community and the general public in recent years, following groundbreaking advances in generative modeling.
The booming of those generative AI techniques brings numerous advantages and conveniences but also raises heightened concerns about the potential malicious usage of their generated fake data, especially within the context of identifiable human faces. 
Our primary motivation in this work is to catch up with the rapidly advancing generative techniques for building an effective protective shield against the potential misuse of generative AI, via a \emph{more comprehensive dataset} and \emph{less biased benchmark protocol}.

\begin{figure*}[t]
    \centering
    \includegraphics[width=1.0\textwidth]{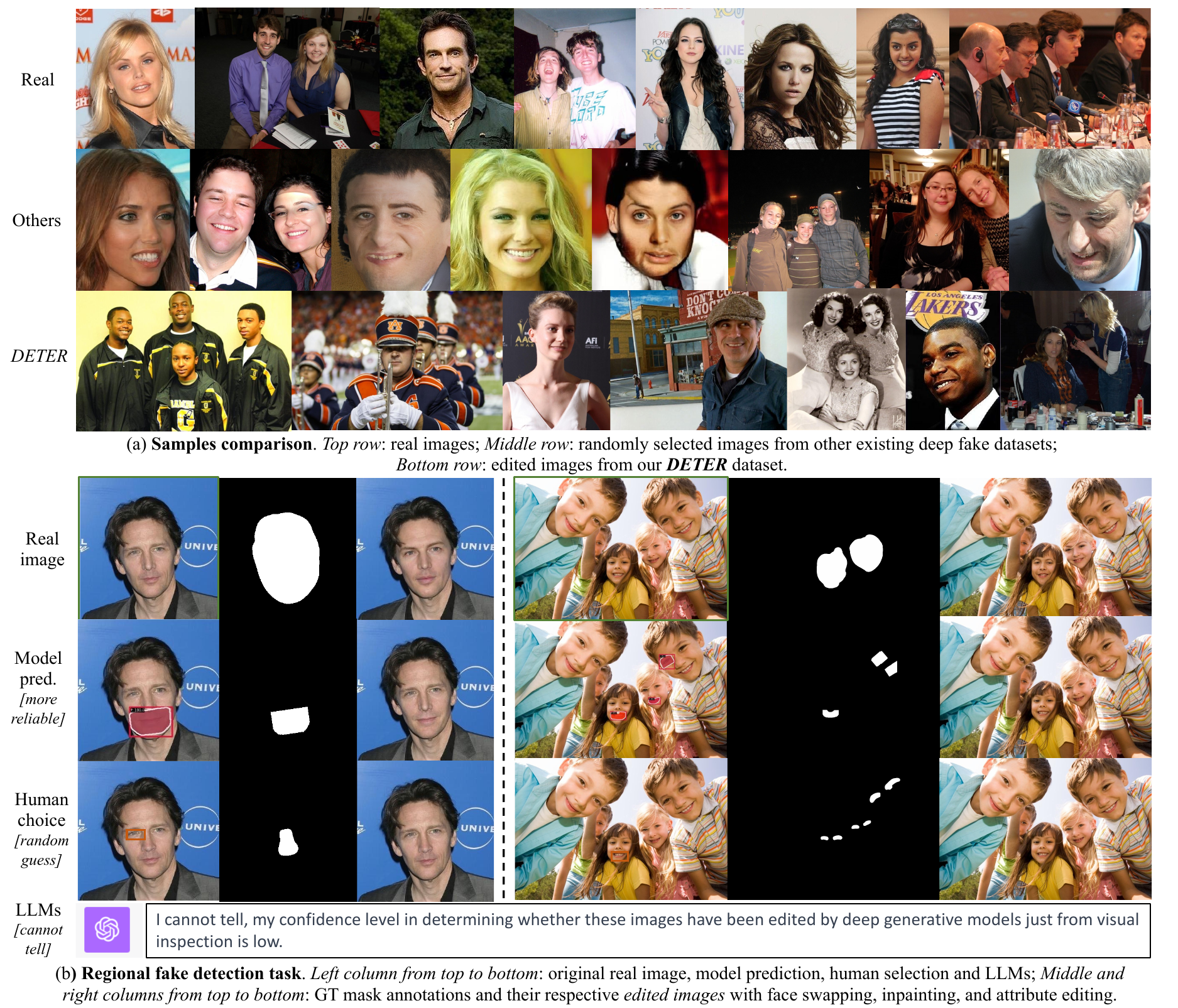}
    \caption{\textbf{In this work, we introduce \emph{DETER} dataset for detecting edited regions manipulated by state-of-the-art generative models.} By formalizing the problem as a regional detection task, models can achieve much better performance than human evaluators and popular Large Language Models (LLMs) such as GPT-4~\cite{OpenAI_ChatGPT}.}
    \label{fig:teaser}
    \vspace{-0.1in}
\end{figure*}

\emph{Existing deep fake datasets are lagging behind in terms of the quality of generators and editing operations during construction.}
Failure to catch up with the fast-developing generative fields in the dataset construction front sets the first gap between the generative and detection teams.
In the upstream generative architecture area, the Diffusion Models (DMs)~\cite{sohl2015dpm_thermo,ho2020dpm,theo-diff2} are replacing Generative Adversarial Nets (GANs)~\cite{gan,karras2017progressive,gal2022stylegan} and become the new state-of-the-art generative models by achieving impressive performance in data generation for images~\cite{stablediff,dhariwal2021diff-img1,ho2022diff-img2,song2020ddim,ramesh2022dalle2,ho2022imagen}, audio~\cite{kong2020diffwave,zhu2022discrete,mittal2021symbolic-diff,lee2021nu}, and videos~\cite{ho2022video,singer2022make}.
On the downstream front, where generative models are leveraged for versatile and fine-grained applications like data editing and customization, both GANs-based~\cite{liu2023fine,yildirim2023diverse,li2022mat,pan2023drag} and DMs-based frameworks~\cite{zhu2023boundary,kim2022diffusionclip,liu2023flowgrad,ruiz2023dreambooth,yang2023diffusion} continue to share an equal footing.
However, as listed in Tab.~\ref{tab:datasets}, none of the existing deep fake datasets~\cite{rossler2018faceforensics,rossler2019faceforensics++,li2020celeb,zi2020wilddeepfake,korshunov2018deepfakes,yang2019exposing,dolhansky2019deepfake,jiang2020deeperforensics1,he2021forgerynet,le2021openforensics} has yet incorporated Diffusion Models as the manipulation sources.
In addition, most manipulations focus on face swapping, with some of them also including attribute editing.
While these used to be representative forgery operations, current generative models can do more than the above.
Particularly, a popular branch of recent works can \emph{achieve very photorealistic and natural effects for image inpainting on arbitrary image regions}~\cite{li2022mat,xia2023diffir,stablediff,lugmayr2022repaint}, which is a novel type of forgery operations that has never been addressed before in the detection side.
It is worth investigating since it presents a different generation mechanism compared to face swapping and attribute editing. While face swapping and attribute editing rely on the information from reference images to ``replace'' the target region of real images, inpainting techniques leverage the generators' ``intrinsic understanding'' of the real images to fill in the missing regions, as illustrated in Fig.~\ref{fig:pipeline}.
\emph{To this end}, our \emph{DETER} presents high-quality images edited by four state-of-the-art generators (at most one year long) based on both GANs and DMs backbones~\cite{xia2023diffir,li2022mat,zhao2023diffswap,liu2023e4s} for three regional editing operations in different granularities (i.e., face swapping, inpainting, and attribute editing, ordering from coarse to fine), each annotated with binary labels, precise masks, and source generators.
We also employ careful post-processing techniques during the dataset construction to ensure that the synthesized image look realistic, including color matching, Poisson fusion and image sharping. To the best of our knowledge, this is the first large-scale regional deep fake dataset with DMs-based generators.
More details about our \emph{DETER} dataset are presented in Sec.~\ref{sec:dataset}, with samples included in Fig.~\ref{fig:teaser}.


\begin{table*}[th]
    \centering
    \scalebox{0.82}{
    \begin{tabular}{l|crrccccccc}
    \toprule
      Datasets   & Format & Real & Fake & GANs & DMs & FaceSwap & Attribute & Inpaint & Multiple faces & Masks  \\ \hline \hline
      DF-TIMIT 18'~\cite{korshunov2018deepfakes} & Videos & 320 & 640 & $\checkmark$& \xmark & $\checkmark$ & \xmark & \xmark & \xmark & \xmark \\
      FaceForensics++ 19'~\cite{rossler2019faceforensics++}& Videos & 1,000 & 5,000 & \xmark & \xmark & $\checkmark$ & \xmark & \xmark & \xmark & $\checkmark$ \\
      Celeb-DF 20'~\cite{li2020celeb} & Videos & 590 & 5,639 & $\checkmark$ & \xmark & $\checkmark$ & \xmark & \xmark & \xmark & \xmark \\
      DFFD 20'~\cite{dang2020detection} & Videos & 1,000 & 3,000 & $\checkmark$ & \xmark & $\checkmark$ & $\checkmark$ & \xmark & \xmark & $\checkmark$\\
      DFDC 20'~\cite{dolhansky2020dfdc} & Videos & 23,564 & 104,500 & $\checkmark$ & \xmark & $\checkmark$ & \xmark & \xmark & \xmark & \xmark\\
      ForgeryNet 21'~\cite{he2021forgerynet} & Videos & 99,630 & 121,617 & $\checkmark$ & \xmark & $\checkmark$ & $\checkmark$ & \xmark & $\checkmark$ & $\checkmark$ \\
      DF-Platter 23'~\cite{narayan2023df}& Videos & 764 & 132,496 & $\checkmark$ & \xmark & $\checkmark$ & \xmark & \xmark & $\checkmark$ & \xmark \\
        SwapMe/FaceSwap 17'~\cite{zhou2017two} & Images & 4,600 & 2,010 &  \xmark & \xmark & $\checkmark$ & \xmark & \xmark & \xmark & \xmark     \\
      OpenForensics 21'~\cite{le2021openforensics} & Images & 45,473 &  115,325 & $\checkmark$ & \xmark & $\checkmark$ & \xmark & \xmark & $\checkmark$ & $\checkmark$\\
      DGM$^4$ 23'~\cite{shao2023detecting} & Images\&Texts & 77,426 &  152,574 &  $\checkmark$ & \xmark & $\checkmark$ & $\checkmark$ & \xmark & \xmark & \xmark \\ \hline
      DETER (Ours) & Images & 38,996 & 300,000 & $\checkmark$ & $\checkmark$ & $\checkmark$ & $\checkmark$ & $\checkmark$& $\checkmark$ & $\checkmark$ \\

         \bottomrule
    \end{tabular}}
\caption{\textbf{Comparison of basic dataset statistics for deep fake detection.} We list the existing popular deep fake datasets in time ordering, with their scales, generators and editing operations. Most existing popular deep fake datasets are video-based, several recent image datasets edit face images with the swapping operation. \emph{DETER} includes the state-of-the-art GANs and DMs-based generators with diverse editing operations and annotations.}
\vspace{-0.1in}
    \label{tab:datasets}
\end{table*}

The quality of our dataset is further guaranteed by extensive \emph{Institutional Review Board (IRB) approved} human studies in two layouts, from which we show solid superior visual quality compared to existing deep fake datasets~\cite{shao2022detecting,shao2023detecting,le2021openforensics}, as well as the challenge for humans to detect regional manipulations.
In addition, as Large Language Models (LLMs) can now achieve superior or similar performance to humans in several existing multimodal tasks such as image captioning~\cite{fei2023transferable}, we also evaluate the deep fake detection on the state-of-the-art LLMs (i.e., GPT-4~\cite{OpenAI_ChatGPT}) by proper prompt engineering.
Details about human studies and LLMs evaluations are included in Sec.~\ref{sec:human}.

\emph{In addition, the current image fake detection benchmark suites tend to introduce spurious patterns in regional detection tasks, which makes the existing detection methods achieve seemingly good performance but biased with an undesired high false alarm rate.}
Binary classification formulation~\cite{wang2023dire,corvi2023detection,ricker2022towards} where a detector classifies the whole image as being ``real'' or ``fake'' is an over-simplified situation compared to the real-life malicious scenarios where often part of the real images are modified without changing the personal identity as shown in Fig.~\ref{fig:teaser}.
As an intuitive step forward, OpenForensics~\cite{le2021openforensics} is the first image dataset that introduces fake regional detection and segmentation benchmark tasks.
However, despite its efforts to bring the fake detection studies closer to a more fine-grained setup, there is a critical gap to fulfill before building reliable fake detection models: the \emph{spurious correlation} problem.
Specifically, with a deeper dive investigation into the current benchmark designs, we note that due to the simplicity of the forgery operation, current detection and segmentation models tend to capture the spurious correlations. 
For example, simplicity of face swapping can cause detection models to predict certain parts of images such as faces as fake regions because they are most frequently seen in the training data), instead of learning the true generative patterns.
Moreover, existing evaluation protocols with classic metrics such as the Average Precision (APs) \emph{fail to reflect} such undesired backdoor shortcut. 
The existing detection and segmentation models~\cite{girshick2015fast,he2017mask} thus can achieve seemingly good performance but are often biased with a high false alarm rate as verified and explained in our extensive experimental results and break-down analysis in Sec.~\ref{sec:detection}.
As mitigation, our inpainting operation in \emph{DETER}, which can be deployed on arbitrary image regions, helps to \emph{decouples the spurious correlations} between certain visual cues (e.g., face shapes) and the actual forgery operations.
In addition, we introduce negative examples (i.e., real images) in the current detection and segmentation setup, accompanied with a \emph{region-based image-level classification accuracy} in the evaluation protocol as an extra assessment criterion to supplement the standard metrics, which helps better reflect the true performance of regional fake detection methods.

To sum up, we believe that \emph{DETER} shall help our community to build more robust and reliable fake detection systems with its main contributions summarized as follows:
\begin{itemize}
    \item We present the \emph{DETER}, the first large-scale dataset that incorporates the SOTA diffusion-based generators for better deterring regional generative manipulations.
    \item We introduce a novel type of regional forgery operation, i.e., inpainting, from the generative modeling side that has been not yet addressed in the detection community.
    \item We note the spurious correlations in the current regional fake detection benchmark suites and propose to mitigate them in \emph{DETER} with inpainting operation in the \emph{dataset construction}, an improved \emph{detection task setup}, as well as extra assessment criterion in the \emph{evaluation protocol}.
\end{itemize}

\section{Related Work}
\label{sec:related}

We present existing literature from domains that are mostly closely related to our work in this section.

\begin{figure*}[t]
    \centering    \includegraphics[width=0.95\textwidth]{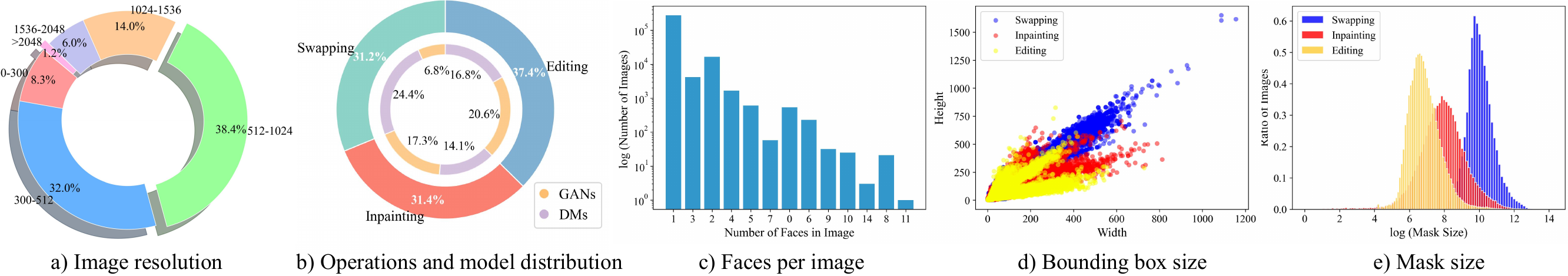}
    \caption{\textbf{Statistical distributions in \emph{DETER}.} Our dataset covers images in diverse resolutions, edited via multiple SOTA generators with different editing operations and versatile mask sizes and shapes.  Best viewed in color with zoom-in.}
    \label{fig:stat}
    \vspace{-0.1in}
\end{figure*}

\noindent \textbf{Deep Fake Datasets.}
Most existing deep fake datasets can be categorized as either video-based~\cite{rossler2018faceforensics,rossler2019faceforensics++,li2020celeb,zi2020wilddeepfake,korshunov2018deepfakes,yang2019exposing,dolhansky2019deepfake,jiang2020deeperforensics1,he2021forgerynet,dang2020detection} or image-based~\cite{le2021openforensics,shao2023detecting,zhou2017two}, as summarized in Table~\ref{tab:datasets}.
All of these datasets provide true or false labels that enable binary classification benchmark tasks, while few of them integrate more fine-grained box or mask-level annotations for fake region detection or segmentation tasks. 
Face swapping with GANs-based generators is the most commonly adopted forgery operation during the construction, with few including attribute editing.
In comparison, \emph{DETER} is the first large-scale dataset that uses the latest state-of-the-art fine-grained methods as generators, and covers editing operations with different granularities. 
Notably, inpainting is a forgery operation that has never been addressed before in deep fake datasets.

\noindent \textbf{Generative Models for Image Manipulations.}
While diffusion models (DMs)~\cite{sohl2015dpm_thermo,ho2020dpm,theo-diff1,theo-diff2,stablediff,ramesh2022dalle2,dhariwal2021diff-img1} are steadily replacing generative adversarial networks (GANs)~\cite{gan,karras2017progressive,gal2022stylegan,xu2018attngan} and have become the dominating method for image synthesis in the past two years, GANs have not yet been entirely supplanted in the downstream side for more fine-grained data manipulation applications such as face swapping and inpainting. 
Among the most recent works that perform fine-grained image manipulations within the past two years~\cite{liu2023fine,yildirim2023diverse,li2022mat,pan2023drag,liu2023e4s,zhao2023diffswap,zhu2023boundary,kwon2022diffusion,lugmayr2022repaint,xia2023diffir}, we carefully select four methods~\cite{liu2023e4s,li2022mat,zhao2023diffswap,xia2023diffir} that cover both GANs and DMs backbones based on their editing quality and versatility as the generators in this work.

\noindent \textbf{Fake Detection Modeling.}
Fake detection methods are closely entangled with available benchmarks and evaluation systems.
Many earlier works~\cite{liu2020global,dang2020detection,li2020face,wang2020cnn,yu2019attributing} tackle the problem against GAN-based generators using Convolutional Neural Networks (CNNs) discriminators and can already achieve very high accuracy (more than 99.9\%) in discerning fake/real images. 
Even the most recent fake detection works that build upon diffusion models~\cite{corvi2023detection,ricker2022towards,wang2023dire} still follow the conventional setting and formalize it as a binary classification problem. 
However, the demand for fake detection methods has gone beyond a true or false label, especially given the more sophisticated generators.
In this work, we formalize the problem as more fine-grained detection and segmentation tasks, but explicitly note that the improvements in terms of the experimental setup are necessary to reflect the true performance.


\section{\emph{DETER} Dataset}
\label{sec:dataset}

In this section, we present the details about our \emph{DETER} dataset and the construction method.

\subsection{Dataset Overview}
\label{subsec:dataset_statistics}

\noindent \textbf{Diverse Real-life Scenarios.}
Among different real image datasets that include humans, we select CelebA~\cite{liu2015faceattributes} and WiderFace~\cite{yang2016wider} as the real human face image sources.
The rationales for the above choices is that CelebA~\cite{liu2015faceattributes} is one of the most widely adopted datasets in the generative modeling area, and WiderFace~\cite{yang2016wider} includes in-the-wild real images that better capture the complex real-life scenarios.
The former includes mostly single-face images while the latter has images with multiple people.
Both datasets are open access to the public under proper license (Creative Common License) for non-commercial research purposes.

\begin{figure}[t]
    \centering
\includegraphics[width=0.47\textwidth]{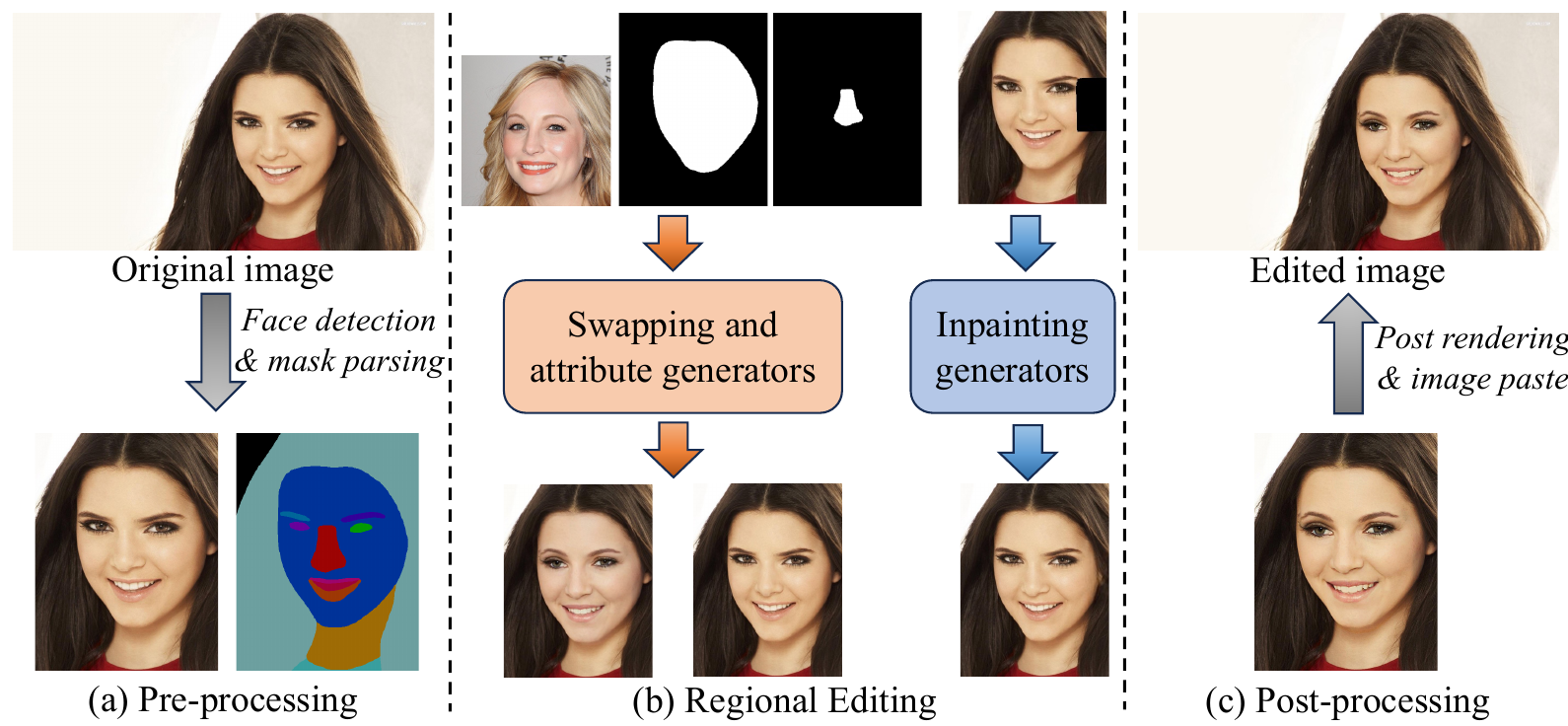}
    \caption{\textbf{Pipeline for our regional fake approach in \emph{DETER} construction.} We first pre-process raw images with face detection and mask parsing techniques~\cite{bulat2017far,yu2018bisenet}. Next, we apply SOTA DMs and GANs-based generators~\cite{liu2023e4s,li2022mat,zhao2023diffswap,xia2023diffir} for different editing operations. Finally, we apply a series of post-rendering techniques and paste the edited regions back to original images, ensuring the high visual quality.}
    \vspace{-0.1in}
    \label{fig:pipeline}
\end{figure}

\noindent \textbf{Editing Operations.}
\emph{DETER} incorporates three image editing operations with different granularities from coarse to fine in terms of editing areas: face swapping, inpainting and attribute editing.
Specifically, \emph{face swapping} involves replacing one person's face in a real image given a reference image (face).
\emph{Inpainting} fills up a missing part of a given image using generative models without other reference images.
\emph{Attribute editing} is similar to \emph{face swapping} but in a finer grain, with replaced regions being facial regions such as eyes, ears and lips.
 These operations include different editing regions grounded by binary masks.
 As illustrated in Fig.~\ref{fig:teaser} and Fig.~\ref{fig:pipeline}, their average editing areas are 31192, 6111, and 1625 measured in pixel numbers, corresponding approximately to squares of 176, 78, and 40, respectively.
 
Particularly, while face swapping and attribute editing are more commonly adopted forgery techniques in existing datasets, inpainting is an \emph{unique} operation in our \emph{DETER}. 
Different from the other conventional techniques, inpainting does not rely on reference images and can be applied to arbitrary regions, which presents a novel and important type of forgery that mitigates the spurious correlations introduced in previous dataset constructions.
As revealed in our experimental results and analysis in Sec.~\ref{sec:detection}, despite having larger editing masks than attribute editing, the inpainted regions are more difficult to detect by current models, \emph{contradicting the common intuition} that ``with more pixels being modified, it should be easier to detect''.



\noindent \textbf{SOTA Generators.}
We adopt four state-of-the-art generative models as the deep generators for dataset construction after having extensively examined and compared their editing quality.
For \emph{face swapping} and \emph{attribute editing}, we adopt the GANs-based E4S~\cite{liu2023e4s} and DMs-based DiffSwap~\cite{zhao2023diffswap}; for \emph{inpainting}, we deploy the GANs-based MAT~\cite{li2022mat} and DMs-based DiffIR~\cite{xia2023diffir} as the manipulation tools.
All the generators are the most recent state-of-the-art methods for fine-grained image manipulations, dating back to only one year at most.

Interestingly, while DMs~\cite{ho2020dpm,theo-diff2,sohl2015dpm_thermo,ho2022diff-img2} are believed to have surpassed GANs~\cite{gan} in unconditional data synthesis, our analysis suggests the current detection methods are more robust against GANs-based generative techniques, as shown in our cross-generator experiments in Sec.~\ref{sec:detection}.


\noindent \textbf{Overall Statistics.}
To sum up, \emph{DETER} presents 300,000 edited images based on 38,996 real images.
The training, validation, and testing splits are partitioned following the 6:1:3 ratio, which includes 180K, 30K, and 90K edited images, respectively. 
We incorporate three editing operations via four SOTA generators.
Our images cover diverse real-life scenarios that includes both single and multiple faces.
Fig.~\ref{fig:stat} summarizes important statistics about our \emph{DETER} dataset with more details can be found in Appendix~\ref{app_sec:dataset}.


\subsection{Regional Editing Approach}
\label{subsec:construction}

Our regional editing approach for the dataset construction is in Fig.~\ref{fig:pipeline}, which mainly include pre-processing, regional editing, and post-processing steps as explained below.

\noindent \textbf{Pre-processing.}
We first run face detection and alignment methods~\cite{bulat2017far} on the real images and parse the detected face to obtain masks with different levels that include the entire face and detailed features such as eyes, lip, and nose~\cite{yu2018bisenet,yu2021bisenet}. 
The selection of editing masks is based on specific operations.
For face swapping and attribute editing, we adopt the face and feature-level masks, respectively. 
As for inpainting, there are two ways to obtain the editing masks: we either dilate the original feature-level masks into arbitrary shapes or randomly pick an image region within the face mask. 
The rationale behind our mask generation mechanism for the inpainting operation is to ensure that it has an editing granularity that is between the face swapping and attribute editing, also further decoupling the spurious correlations between low-level face feature characteristics and the editing operations, increasing the difficulties for the model detection as revealed by our experiments in Sec.~\ref{sec:detection}.

\noindent \textbf{Regional Editing.}
After the pre-processing of raw images and the mask selection, we then proceed to the specific editing step.
As previously mentioned, we use GANs-based E4S~\cite{liu2023e4s} and MAT~\cite{li2022mat}, and DMs-based DiffSwap~\cite{zhao2023diffswap} and DiffIR~\cite{xia2023diffir} as the deep generators. 
Specifically for face swapping and attribute editing, the deep generators also take a reference image as input in addition to the original face image and binary masks. 
In contrast, inpainting models take an image with missing regions grounded by our editing masks as input and output an image completed by generative models, as illustrated in Fig.~\ref{fig:pipeline}.

\noindent \textbf{Post-processing.}
To better ensure the quality of our \emph{DETER} dataset, we apply a series of post-rendering techniques on the output of various deep generators, which include color matching, Poisson fusion, and image sharping. These operations alleviate the boundary effects (i.e., low-level visual image distortions perceivable by human eyes) in conventional forgery construction pipelines and further boost the quality of our dataset.
We then paste the face regions back into the original images to get the final edited images, which strictly ensures that our mask annotations precisely reflect the actual regions manipulated by generative models.

\noindent \textbf{Better Visual Quality and ID Preservation.}
We demonstrate the high quality of our dataset in both qualitative and quantitative assessments. 
On the one hand, as illustrated by the samples in Fig.~\ref{fig:teaser}, the edited images from \emph{DETER} can be hardly detected by bare eyes, which is further confirmed by our human studies in the next section.
On the other hand, our dataset has a lower ID-distance~\cite{yang2023diffusion} score of \textbf{0.30}, compared to the most recent DGM$^4$~\cite{shao2023detecting} dataset that has the same score of \textbf{0.93} based on 10,000 sampled images, indicating \emph{DETER} has the better identity preservation quality.




\section{Human Study and LLMs}
\label{sec:human}

To validate the visual consistency and harmony of the proposed dataset, we perform human studies and LLMs-based evaluations with the state-of-the-art GPT-4 from OpenAI~\cite{OpenAI_ChatGPT}.
The human studies and data analysis are conducted under appropriate Institutional Review Boards (IRB) approval and regulations.
While humans are usually believed to be the performance upper-bound in various computer vision tasks to ground the model learning such as object recognition and segmentation~\cite{zhao2019object,minaee2021image}, they become \emph{lower-bound} in fake detection. ChatGPT performs even worse than human in this case, which further confirms the quality of our \emph{DETER}, as well as the great potential and necessity of model assistance when deploying responsible Generative AI in real life.


\subsection{General Quality Assessment}

In the first layout of human studies, we investigate the human performance on the general fake detection, which resembles the conventional binary image classification problem, similar to previous works~\cite{rossler2019faceforensics++,liu2020global,le2021openforensics}.

Specifically, we prepare 400 image triplets, with each including two real images and one edited image, and ask the human evaluators to pick the fake one. 
In addition to the above three image options, we also include a supplementary \emph{``I am not sure''} option, which allows the evaluators to forfeit instead of forcing them to make a choice when it comes to hard samples.
Among 400 edited images, half of them are randomly selected from our \emph{DETER}, with another half equally sampled from existing deep fake sources including SeqDeepFake~\cite{shao2022detecting}, DGM$^4$~\cite{shao2023detecting}, OpenForensics~\cite{le2021openforensics}, and DDPMs~\cite{ho2020dpm}.
We report the distribution of picks, as well as the detection rate conditioned on correct picks in Tab.~\ref{tab:human_tab1}.
Given \emph{an equal population} of fake images, the detection rate on \emph{DETER} is \textbf{20.4\%} lower than the ensemble of other fake sources, which demonstrates the high visual quality of our edited images.



\begin{table}[ht]
    \centering
    \scalebox{0.71}{
    \begin{tabular}{l|cccc|c}
    \toprule
      Choices   & Real & Others datasets & \emph{DETER} & Unsure & Total  \\ \hline \hline
       Human picks & 38.3\% & 23.7\%  &  15.7\% & 22.3\% & 100\%\\
       Human detection rate & - & 60.2\% & \textbf{39.8\%} &- & 100\% \\
       LLMs picks & 0\% & 3\% & 2\% & 95\% &100\%\\
       LLMs detection rate & -& 60\% & \textbf{40\%} & -&100\% \\
        \bottomrule
    \end{tabular}}
    \caption{\textbf{User-study and LLMs results for general quality assessment.} ``Human picks'' indicates humans/LLMs prefer marking the candidate as the fake image, while ``detection rate'' represents the pick proportion over all picked fake images.
    Given an equal population, the detection rate is much lower on \emph{DETER} compared to the ensemble of other fake sources, demonstrating \emph{DETER} has \emph{better visual quality} more difficult to be detected by humans.}
    \vspace{-0.1in}
    \label{tab:human_tab1}
\end{table}


\vspace{-0.1in}
\subsection{Regional Fake Detection}
We conduct another more fine-grained layout of human evaluations for regional fake detection using another 100 triplets.
In this case, each image triplet comprises the same edited image from \emph{DETER}, with each one grounded in different regions. One of these regions represents the ground truth, while the other two are randomly selected as distractors.
Similar to the first layout, we ask the human evaluators to pick among the three candidate regions or choose to forfeit by picking the \emph{``I am not sure''} option.
The results in Tab~\ref{tab:human_tab2} show that this is a more challenging task for humans, with the rate to pick the GT regions close to a random guess.

\begin{table}[ht]
    \centering
    \scalebox{0.85}{
    \begin{tabular}{l|ccc|c}
    \toprule
      Choices   &  Random regions & GT & Unsure & Total\\ \hline \hline
       Human picks  & 59.0\% & 30.3\% & 11.7 \% & 100\%\\
       LLMs picks & 0\% & 0\% & 100\% & 100\% \\
        \bottomrule
    \end{tabular}}
    \caption{\textbf{User-study and LLMs results on regional fake selection.} Picking the edited region is a more challenging task for human evaluators, and the rate for picking the GT regions is similar to a random guess. }
    \vspace{-0.1in}
    \label{tab:human_tab2}
\end{table}


\subsection{LLMs}

To comprehensively assess the quality of the \emph{DETER}, we also adopt GPT-4~\cite{OpenAI_ChatGPT}, the state-of-the-art LLMs that can process multimodal information including images and texts, for evaluating the fake detection performance.
We follow proper prompt tuning to setup the similar evaluation process as the human studies for general quality assessment and regional fake detection.
The results are integrated into Tab.~\ref{tab:human_tab1} and Tab.~\ref{tab:human_tab2} based on our 100 queries for each evaluation task.
Based on our evaluations, GPT-4 tends to give an uncertain answer, by picking the option \emph{``I am not sure''}.

\section{Benchmark for Regional Fake Detection}
\label{sec:detection}

In this section, we present our benchmark suites based on the \emph{DETER} dataset for the regional deep fake detection task.

\begin{table*}[t]
\centering
\scalebox{0.65}{
\begin{tabular}{l|c|ccccccccccccccc}
\toprule
Methods                     & \multicolumn{1}{c|}{}           & \multicolumn{3}{c|}{\begin{tabular}[c]{@{}c@{}}Classification\\ (image-level)\end{tabular}} & \multicolumn{9}{c|}{\begin{tabular}[c]{@{}c@{}}Object Detection\\ (box-level)\end{tabular}}                                                                                                                                 & \multicolumn{3}{c}{\begin{tabular}[c]{@{}c@{}}Instance Segmentation\\ (mask-level)\end{tabular}} \\ \hline
                            &  & \multicolumn{1}{c|}{Swap}  & \multicolumn{1}{c|}{Inpaint}  & \multicolumn{1}{c|}{Attribute} & \multicolumn{3}{c|}{Swap}                                               & \multicolumn{3}{c|}{Inpaint}                                            & \multicolumn{3}{c|}{Attribute}                                          & \multicolumn{1}{c|}{Swap}           & \multicolumn{1}{c|}{Inpaint}          & Attribute          \\ \cline{2-17} 
                            & \multicolumn{1}{c|}{Setup}    & \multicolumn{3}{c|}{Accuracy}                                                               & Precision             & Recall                & \multicolumn{1}{c|}{AP} & Precision             & Recall                & \multicolumn{1}{c|}{AP} & Precision             & Recall                & \multicolumn{1}{c|}{AP} & \multicolumn{3}{c}{Mask AP}                                                                      \\ \hline \hline

MaskR-CNN 17'~\cite{he2017mask} & \multirow{6}{*}{Conventional}   & 0.51         & 0.43            & 0.41             & 0.25      & \textbf{0.97}   & \textbf{0.97} & 0.24      & 0.92   & 0.86 & 0.35      & 0.95   & 0.87 & \textbf{0.96}           & 0.85              & 0.87               \\

Yolact 19'~\cite{bolya2019yolact}                 &          &  0.52            & 0.45               &   0.45      & \underline{0.08}          & \textbf{0.97}       & 0.96     & \underline{0.06}          & 0.89       & 0.77     & \underline{0.10}         & 0.91       &  \underline{0.77}    & 0.96               & 0.75                  &  \underline{0.77} 
   \\

Mask2Former 22'~\cite{cheng2022masked}                 &          &  0.47           & 0.42                 &  \underline{0.40}             & 0.20         & \textbf{0.97}     & \underline{0.95}   &  0.20      & 0.88       & \underline{0.73}    & 0.31         &  0.92      &  0.84    &  \underline{0.95}             & \underline{0.73}              &   0.84
           \\

FasterR-CNN 15'~\cite{girshick2015fast}                &          &  0.53    & 0.43      & 0.41   & 0.27  &  \textbf{0.97} & \textbf{0.97}  &   0.25    &  0.90   &  0.83 &   0.37     &  0.93      & 0.85    &         -      &    -              &   -
           \\

YOLOX 21'~\cite{ge2021yolox}                &               &  0.54           &  0.51               &   0.52            &   0.29        &  \underline{0.96}     &  0.96   &    0.30     & 0.91        & 0.80      &    0.43      &  0.93      &   0.86   &  -             &   -               &   -
           \\ 
 
\
DINO 22'~\cite{zhang2022dino}                &              &         \underline{0.44}    & \underline{0.38}                 &  0.41  &        0.11    & \textbf{0.97}          &   0.96    & 0.11    & \textbf{0.93}        &   0.84     &   0.19   & \textbf{0.96}         &     0.87   &    -                &   -               &  - 
           \\ 
\hline

MaskR-CNN 17'~\cite{he2017mask} &  \multirow{6}{*}{Improved}    & 0.75      &   0.68          &  0.64            & 0.45      &  \textbf{0.97}  &  0.96 &  0.41     & 0.91   & \textbf{0.88} &  0.53     & 0.93   & 0.89 &  0.96          &    \textbf{0.88}          &   \textbf{0.89}             \\  
Yolact 19'~\cite{bolya2019yolact} &      &  0.85        &  0.78           &    0.74          &  0.47     &  \textbf{0.97}  & 0.96 & 0.35      & 0.88   & 0.85 & 0.45      & \underline{0.88}   &  0.83 & 0.96           &   0.83           &   0.83             \\  
 Mask2Former 22'~\cite{cheng2022masked}       &      & 0.78         &   0.70          &     0.65      & 0.44      & \textbf{0.97}   & 0.96 & 0.37      &  \underline{0.87}  &0.83  &  0.48     & 0.90   & 0.84 & 0.96     & 0.83             &   0.84             \\  
FasterR-CNN 15'~\cite{girshick2015fast}       &      &    0.77      &  0.69           & 0.65     &  0.50     & \textbf{0.97}   & 0.96 &  0.43     &  0.89  & 0.86 &  0.55     & 0.91   & 0.87 &   -         & -             &   -             \\
YOLOX 21'~\cite{ge2021yolox}       &      &    \textbf{0.92}  &  \textbf{0.86}      &  \textbf{0.82}       &    \textbf{0.78}  &  \underline{0.96}  & \underline{0.95} & \textbf{0.68}      & 0.90   & \textbf{0.88} & \textbf{0.74}      & \underline{0.88}   & 0.85 &    -        &       -       &         -       \\ 
  DINO 22'~\cite{zhang2022dino}       &      & 0.74         &   0.67          &  0.67            &   0.28    & \textbf{0.97}   &  \textbf{0.97} & 0.22      & \textbf{0.93}   & \textbf{0.88} &   0.36    &  \textbf{0.96}  & \textbf{0.92} &        -    &          -    &       -         \\  \hline
  
\bottomrule
\end{tabular}}
\caption{\textbf{Quantitative evaluation results for regional fake detection under \emph{(C)}onventional (i.e., training w/o negative image samples) and \emph{(I)}mporved (i.e., training with negative image samples) settings.} We only report the scores calculated with IoU=0.5 in the main paper due to space limit, with more results in Appendix~\ref{app_sec:detection}. All metrics are the higher the better; \textbf{best} and \underline{worst} results are marked in \textbf{bold} and \underline{underlined}, respectively. Note that the \emph{region-based image-level classification accuracy} is an extra metric in our evaluation protocols that explicitly reflects the \emph{image-level} false alarm rate within the formulation of regional detection and segmentation.}
\vspace{-0.1in}
\label{tab:results}
\end{table*}

\subsection{Improved Experimental Setup}

As we have previously mentioned, current evaluation benchmark suites for regional fake detection tend to introduce spurious correlations during the dataset construction, which leads to biased seemingly good performance under the conventional detection and segmentation task setup and evaluation protocols.

\noindent \textbf{Testing Setting Closer to Practice.}
The testing setup aligns with our improved training designs, in which we incorporate another 90K unedited real images, and each operation task comprises 30K distinct images.
The motivation for our testing design again is to create a setup closer to the practical situations where a large population of images should be free of generative manipulations. 

\noindent \textbf{Conventional and Improved Training Settings.}
In the conventional object detection and instance segmentation training setting, the model learns to distinguish positive and negative regions \emph{within the same image}.
However, relying solely on intrinsic features for self-comparison introduces a strong prior: the models tend to \emph{assume there should always be the presence of target regions} given an image.
This bias is even further amplified in the case of the regional fake detection problem due to relatively fixed edited region patterns and generators, but it contradicts the real-life scenario where a large number of images on the Internet should be real images without any generative manipulations.
To this end, we investigate two training settings in our experiments: the conventional setup with no image-level negative samples (i.e., all the training images are from our \emph{DETER} training split, each including at least one positive edited region), and an improved setting with image-level negative samples (i.e., the mixture of our training split and another 140K unseen real images without any manipulations).

\noindent \textbf{Improved Evaluation Protocols.}
Our evaluation protocols include classic metrics for detection and segmentation tasks, as well as an extra \emph{region-based image-level classification accuracy}.
For classic evaluation metrics, we adopt Precision, Recall, the standard COCO-style Average Precision (AP)~\cite{lin2014microsoft} for box-level detection, and the Segmentation AP for instance segmentation.

The objective of the region-based image-level classification accuracy is to \emph{reflect the image-level false alarm rate}, as supplementary to Precision which reveals the box-level false positives. 
The detection and segmentation models trained in our improved setup predict regions that are believed to be edited by generative models. 
During inference, we first count the detected boxes with an IoU greater than 0.5 with the ground truth ones as positive regions. If there are \emph{no missed detected regions or false positives} in the given image, we then consider it as correctly classified.
Note this is \emph{different from} the conventional image binary classification, but rather an extra metric introduced \emph{within the formulation of detection and segmentation tasks}.

\subsection{Experimental Details}
\noindent \textbf{Baseline Methods.}
We experiment with six detection and segmentation models covering the most classic to the state-of-the-art methods: Mask R-CNN~\cite{he2017mask}, Yolact~\cite{bolya2019yolact}, Mask2Former~\cite{cheng2022masked}, Faster R-CNN~\cite{girshick2015fast}, YOLOX~\cite{ge2021yolox}, and DINO~\cite{zhang2022dino}. Among these methods, Mask R-CNN~\cite{he2017mask} and Faster R-CNN~\cite{girshick2015fast} stand out as well-known convolutional-based two-stage methods, providing a reliable baseline. 
Mask2Former~\cite{cheng2022masked} and DINO~\cite{zhang2022dino} build upon the success of DETR~\cite{carion2020end}, utilizing the transformer-based architecture to model detection and instance segmentation as a direct set prediction. 
The remaining methods are single-stage and aim at real-time performance.

\noindent \textbf{Implementation Details.}
All the methods used ResNet50~\cite{he2016deep} as the backbone for a fair comparison, except for YOLOX~\cite{ge2021yolox}, which utilized DarkNet53~\cite{redmon2018yolov3}. The models were initialized with COCO pretrained weights to enhance performance. We adhered to default settings with slight modifications in epochs and trained the models on 8 Nvidia RTX 4090.
Specifically, for the improved training setting, we do not skip the real images with no forgery regions, but use them as abundant negative samples to update the region proposal networks or classifiers in contrast to the default training where data samples with no foreground bounding boxes usually are skipped.


\subsection{Evaluation Results and Analysis}
\label{subsec:automatic_results}

We present break-down experimental results and analysis below, with more details can be found in Appendix~\ref{app_sec:detection}.

\begin{table}[t]
\centering
\ra{0.98}
\scalebox{0.8}{
\begin{tabular}{cccccccccc}
\hline
    \multicolumn{1}{c}{\multirow{3}{*}{\diagbox[trim=l,height=2.1\line]%
    {\\ \\Train}{Test\\ \\}} }& \multicolumn{3}{c}{Inpaint} & \multicolumn{3}{c}{Attribute}\\  &\multirow{1}{*}{Precision} & \multirow{1}{*}{Recall} & \multirow{1}{*}{AP} & \multirow{1}{*}{Precision}& \multirow{1}{*}{Recall} & \multirow{1}{*}{AP} \\ \\ \hline
     Inpaint  & 0.66& 0.92& 0.91 & 0.47 & 0.47 &0.39\\
     Attribute & 0.07 & 0.23 & 0.08 & 0.50 & 0.95 & 0.90\\                       
\bottomrule
\end{tabular}}
    \caption{\textbf{Quantitative results in terms of \emph{Inpainting} and \emph{Attribute editing} data in cross-domain experiments with Mask R-CNN~\cite{he2017mask}.} Scores are calculated with IoU=0.5. Models trained with inpainting data achieve better cross-domain performance.}
    \vspace{-0.1in}
    \label{tab:cross-domain}
\end{table}

\noindent \textbf{Spurious Correlations and Inpainting.}
The editing regions for face swapping, attribute editing, and inpainting operations are approximately squares of 176, 78, and 40, respectively.
While the detection difficulties are seemingly related to the area of edited regions by intuition, i.e., larger areas of modification tend to be easier to detect, we observe that this does not hold for current detection and segmentation models as shown in Tab.~\ref{tab:results}.
Specifically, we note the edited regions with \emph{inpainting} are consistently more difficult to predict compared to both \emph{face swapping} and \emph{attribute editing}. For example, the precision on inpainting data is on average 0.11 lower (i.e., 0.30 versus 0.41) than that of attribute editing across all models.
The operation-wise difficulty variance further validates our initial claim on the spurious correlations introduced in the dataset construction stage with oversimplified editing types.
Our proposed \emph{DETER} dataset seeks to mitigate the above by integrating \emph{inpainting} to diversify the editing regions and shapes.

\noindent \textbf{Generalization Ability of Inpainting.}
We conduct cross-domain experiments to study the generalization ability of different editing operations. 
As shown in Tab.~\ref{tab:cross-domain}, the model performs much better in in-domain testing (training and testing on the same editing operation) and performs worse in the cross-domain case. 
We also observe the model trained on the inpainting data has better cross-domain generalization performance compared to the one trained on attribute-edited data. The main reason is that the flexible inpainting operation in \emph{DETER} can be applied on arbitrary face parts, and thus, the model captures the better intrinsic difference between real and manipulated regions, rather than just memorizing the position prior/bias in the training data.
As a result, the model trained on inpainting data has a precision of 0.47 on attribute-edited test samples, similar to the in-domain test precision of 0.50.
Our take-away message here is that regional fake detection models should consider the inpainting training data to avoid spurious correlation.

\begin{figure}[t]
    \centering
    \includegraphics[width=0.47\textwidth]{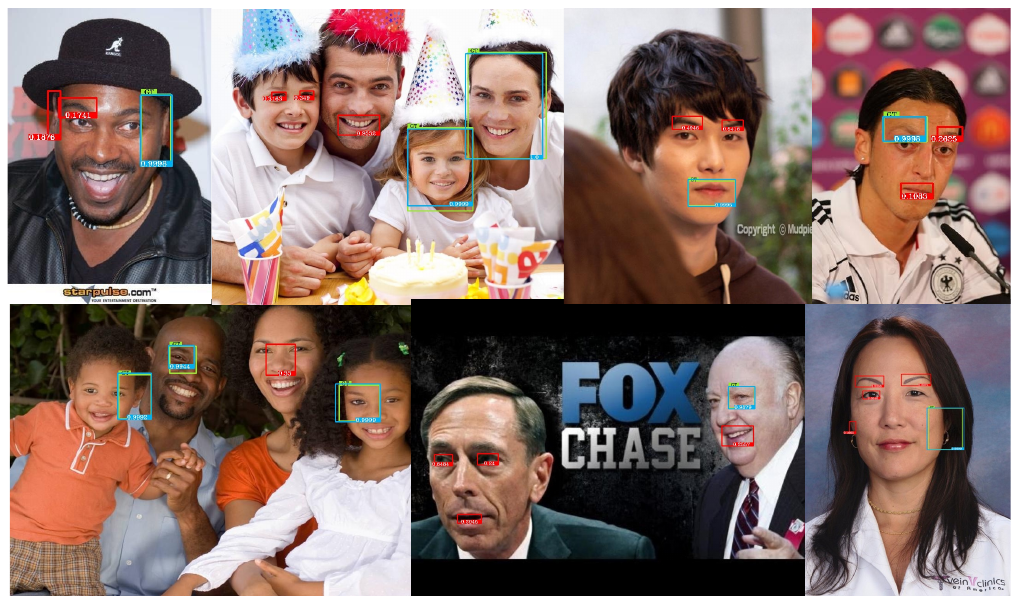}
    \caption{\textbf{Qualitative results of regional fake detection from Mask R-CNN~\cite{he2017mask}.} GT, correct predictions, and false positives are annotated in green, blue, and red boxes, respectively. Current models induce a relatively high false alarm rate. Best viewed in color.}
    \label{fig:qualitative}
    \vspace{-0.11in}
\end{figure}

\noindent \textbf{Image-level and Region-level False Alarms.}
The comparisons among various metrics further reveal the high false alarm rate across existing detection and segmentation methods. 
Particularly, the models tend to achieve very high recall (e.g., greater than 0.9) but low precision (e.g., lower than 0.3) in the conventional setup.
This recall-precision contrast indicates that the models' predictions involve a large number of real regions that have been predicted as fake, as shown in Fig.~\ref{fig:qualitative}.
The same issue is further supported by our \emph{region-based image classification accuracy}, through which we find a lot of real images are classified as edited, resulting in low classification accuracies.
This is undesired when deploying a reliable regional fake detection system in practice, where most images on the Internet should still be free of generative manipulations. 
We believe that a reliable regional fake detection system should ideally \emph{detect edited regions without giving many false alarms}.

\noindent \textbf{Improved Setup with Negative Samples.}
Another dimension of our break-down analysis focuses on the improved task setup with mixed real images in training.
Tab.~\ref{tab:results} also include the evaluation results obtained under both conventional training and improved training setup.
Our improved setup \emph{significantly} boost the classification accuracy and precision by \emph{more than 20\%} across operations and methods, demonstrating its effectiveness.


\noindent \textbf{GANs vs. DMs Generators.}
As a further deep-dive analysis, we also conduct cross-domain experiments on the generative models, with results shown in Tab.~\ref{tab:gan_diff}.
We report the Precision, Recall, and AP scores under the \emph{inpainting} operation task trained with the conventional setting as an illustration example (more generator-based cross-domain results in Appendix~\ref{app_sec:detection}). 
We observe that detection models trained with the GANs-based generators can generalize well to the DMs-based testing images, while the inverse setting induces a non-trivial performance drop. 
The above experiments suggest that the GANs-based generators include more robust features that are perceivable by detection and segmentation models, which have not yet been explicitly revealed in the generative modeling area.

\begin{table}[t]
\centering
\ra{0.98}
\scalebox{0.8}{
\begin{tabular}{cccccccccc}
\hline
    \multicolumn{1}{c}{\multirow{3}{*}{\diagbox[trim=l,height=2.1\line]%
    {\\ \\Train}{Test\\ \\}} }& \multicolumn{3}{c}{GANs} & \multicolumn{3}{c}{DMs}\\  &\multirow{1}{*}{Precision} & \multirow{1}{*}{Recall} & \multirow{1}{*}{AP} & \multirow{1}{*}{Precision}& \multirow{1}{*}{Recall} & \multirow{1}{*}{AP} \\ \\ \hline
     GANs    & 0.48 & 0.90 & 0.87 & 0.48 & 0.91 & 0.88 \\
     DMs & 0.38 & 0.76 & 0.71 & 0.53 & 0.92 & 0.89\\                                   
\bottomrule
\end{tabular}}
\caption{\textbf{Quantitative results in terms of GANs-based and DMs-based generators in cross-domain experiments with Mask R-CNN~\cite{he2017mask}.} The scores are calculated with IoU=0.5. Models trained with GANs-based generators can achieve \emph{equally good performance} when tested with DMs-based testing images. In contrast, DMs-based training data \emph{induce performance flip when tested in GANs-based testing images.}}
\vspace{-0.11in}
\label{tab:gan_diff}
\end{table}

\section{Discussion and Conclusion}
\label{sec:conlusion}

\noindent \textbf{Broader Social Impact.}
We seek to raise awareness of the potential malicious impact of generative models and support future research on building effective and reliable detection systems, by introducing an up-to-date, high-quality, and large-scale comprehensive image benchmark for deterring regional generative manipulations. 


\noindent \textbf{Ethical Conduct.}
All of the images in \emph{DETER} are derived from existing public open access datasets under proper license (Creative Common License) for non-commercial research purposes.
The human studies and data analysis are conducted under appropriate Institutional Review Board approval and regulations.
None of sensitive or personally identifiable information is collected during our studies.

\noindent \textbf{Conclusion and Future Directions.}
We introduce our \emph{DETER} dataset for the regional deepfake detection task, featuring an up-to-date, large-scale, high-quality image dataset. 
We ensure the quality of our benchmark to catch up with the fast-developing generative AI techniques,
including SOTA generators, novel forgery operations, deep-dive investigations on current benchmarks and their problematic spurious correlation issues, as well as improved benchmark designs as mitigation.

For future research on the detection methods, we explicitly emphasize the significance of a more comprehensive and less biased evaluation system that reflects the real performance of models, with particular attention on the false alarm rate when deployed in real-life scenarios.

\section*{Acknowledgements}

We appreciate the support from Princeton First Year Fellowship for Amaya Dharmasiri; also the Princeton SEAs Howard B. Wentz, Jr. Junior Faculty Award for Olga Russakovsky.
We thank William Yang, Angelina Wang, and Max Gonzalez Saez-Diez for helpful input to this work.
This work was also supported in part by the National Natural Science Foundation of China under Grant 62372341.

{
    \small
    \bibliographystyle{ieeenat_fullname}
    \bibliography{main}
}

\clearpage
\setcounter{page}{1}
\maketitlesupplementary

We provide additional details about our \emph{DETER} dataset in Sec.~\ref{app_sec:dataset}.
Sec.~\ref{app_sec:humanstudy} describes more details about our human studies.
More experimental results and analysis can be found in Sec.~\ref{app_sec:detection}.

\section{More Details about \emph{DETER}}
\label{app_sec:dataset}

\begin{figure*}[th]
    \centering
    \includegraphics[width=0.98\textwidth]{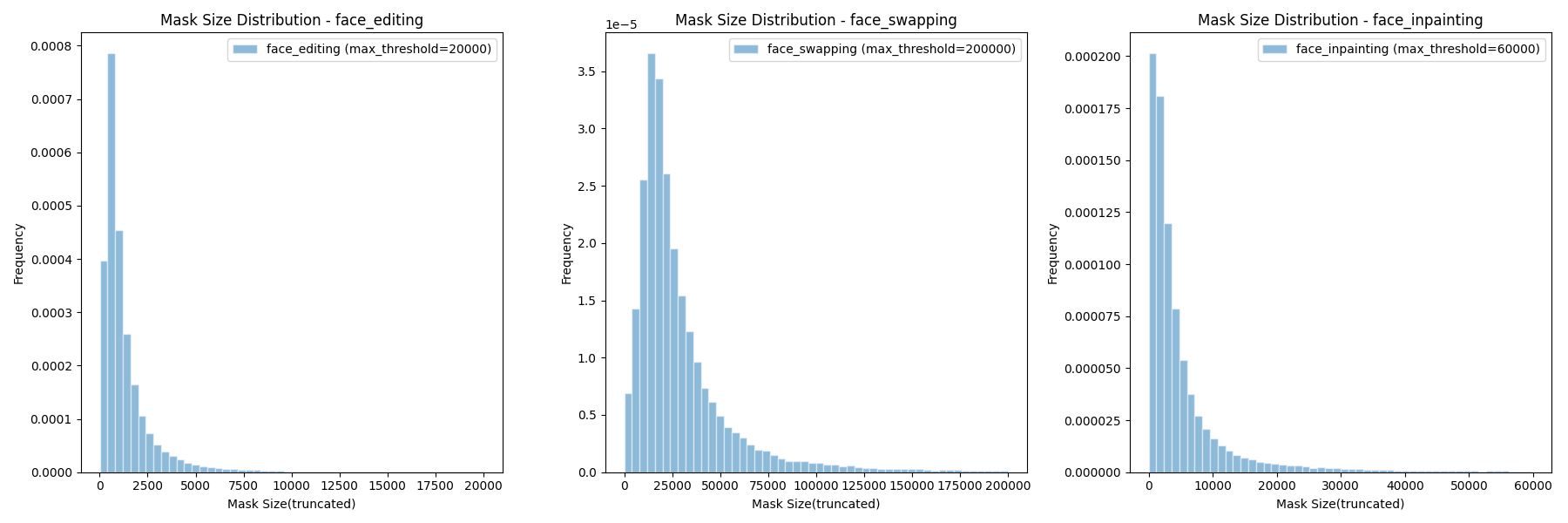}
    \caption{\textbf{Distributions of mask sizes in terms of different manipulation operations.}}
    \label{fig:mask1}
\end{figure*}

\begin{figure*}[th]
    \centering
    \includegraphics[width=0.98\textwidth]{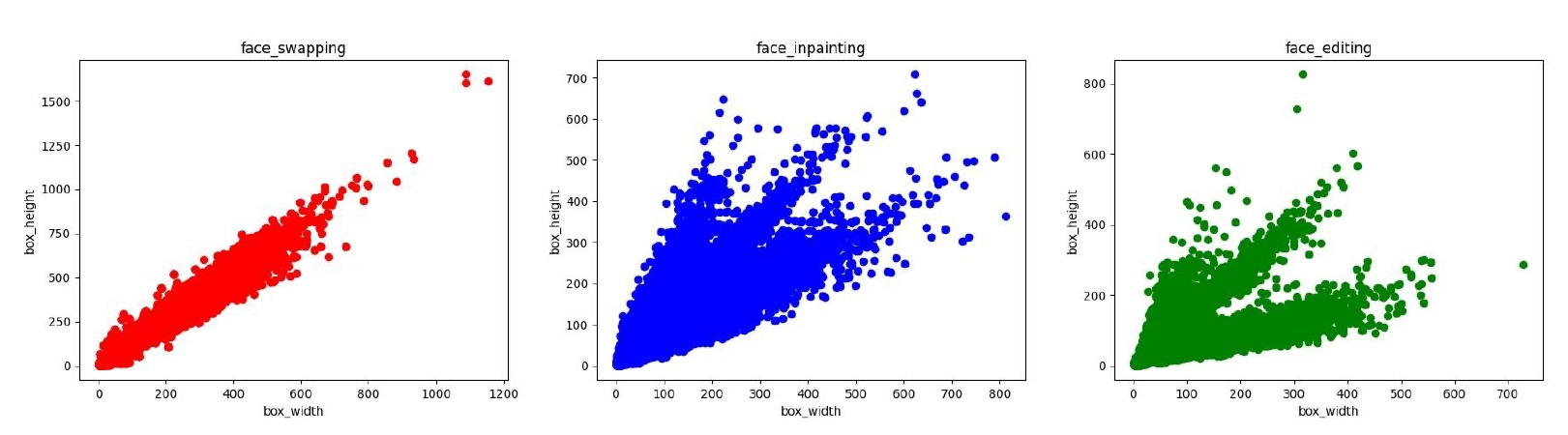}
    \caption{\textbf{Details of box sizes.} \emph{Face swapping} operation has the largest average editing area, followed by \emph{inpainting}, and \emph{attribute editing}.}
    \label{fig:box2}
\end{figure*}

\emph{DETER} includes 300,000 edited images in total, obtained with three editing operations, as described in our main paper.
For face swapping, inpainting, and attribute editing, there are 93636, 94253, and 112111 images, which corresponds to 106673, 114066, and 199958 regional manipulation masks, respectively.
The image resolutions vary based on the real images, from smaller than 300 to greater than 2048.
Fig.~\ref{fig:mask1} and Fig.~\ref{fig:box2} show the distributions of editing masks and their detailed box heights and widths.

\begin{figure*}[th]
    \centering
    \includegraphics[width=0.98\textwidth]{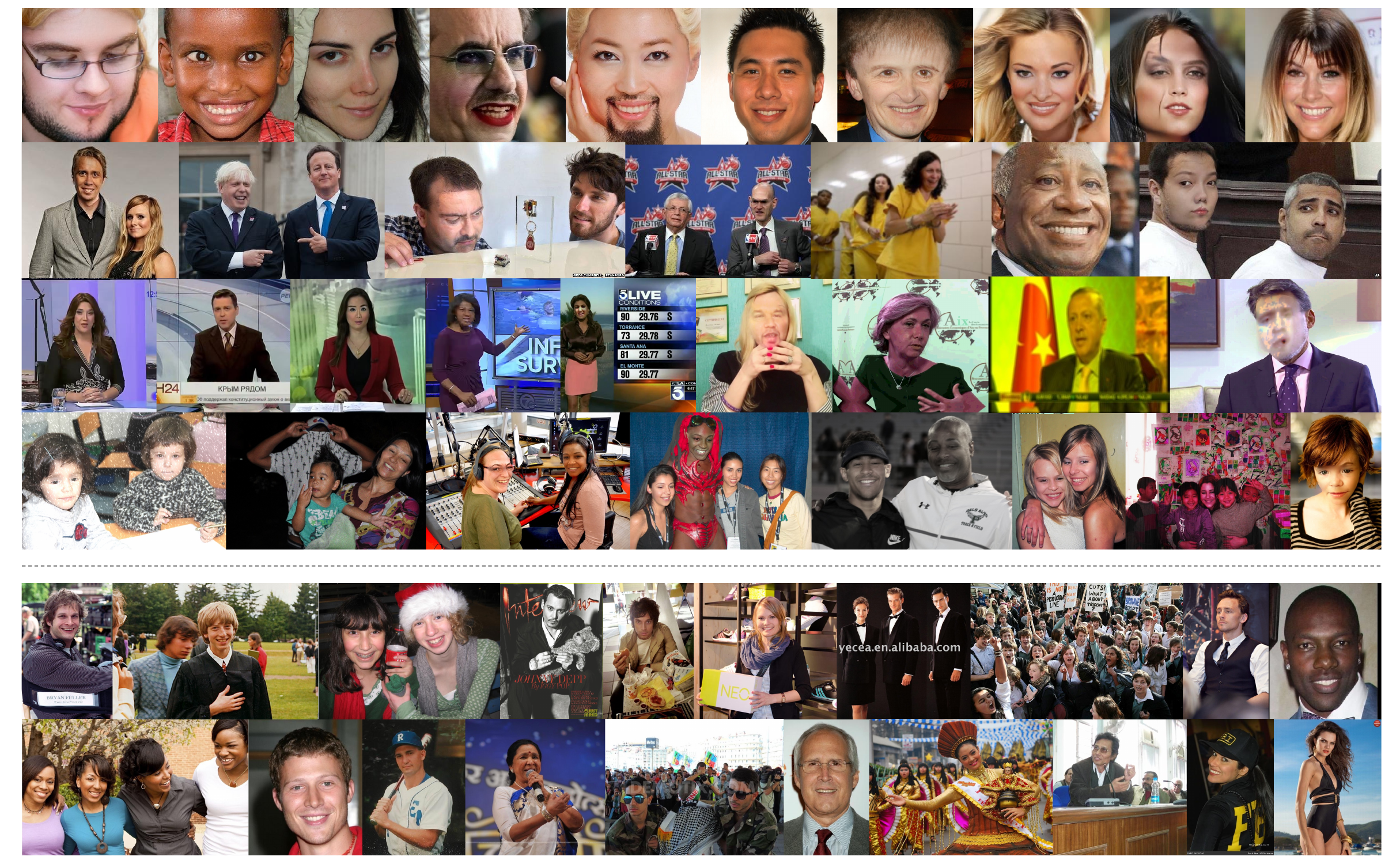}
    \caption{\textbf{More qualitative comparisons among samples from different deep fake datasets.} Image samples in \emph{upper} rows come from existing deep fake datasets: DFFD 20'~\cite{dang2020detection}, SeqDeepFake 22'~\cite{shao2022detecting}, DGM$^4$ 23'~\cite{shao2023detecting}, FaceForensics++ 19'~\cite{rossler2019faceforensics++}, ForgeryNet 21'~\cite{he2021forgerynet}, and OpenForensics 21'~\cite{le2021openforensics}, while the \emph{bottom} rows include samples from our \emph{DETER}.}
    \label{fig:appendix_comparison}
\end{figure*}

Fig.~\ref{fig:appendix_comparison} show more qualitative comparisons between other existing deep fake datasets including DFFD 20'~\cite{dang2020detection}, SeqDeepFake 22'~\cite{shao2022detecting}, DGM$^4$ 23'~\cite{shao2023detecting}, FaceForensics++ 19'~\cite{rossler2019faceforensics++}, ForgeryNet 21'~\cite{he2021forgerynet}, and OpenForensics 21'~\cite{le2021openforensics}.

\section{More Details about Human Studies}
\label{app_sec:humanstudy}

This section describes further details about our human studies.
We organize our human studies in two settings. 
The first task: \textbf{General Quality Assesment} is selecting the fake image from a triplet of 2 real photos and a fake.
This task is aimed at evaluating the difficulty of spotting the fake images generated by our method vs other methods used in existing datasets.
We use human error in selecting the fake image, as a proxy for the difficulty of spotting cues of deepfake generation, hence the realistic quality of the fake image sample. 

The second task: \textbf{Regional Fake Detection} is to select the edited region of a photo. We create samples with a specific facial feature/ region of the face edited or altered using our method. 
Each image triplet for this task involves the same edited image from \emph{DETER} but grounded with different regions, among which one is the ground truth region that has been edited, with the other two untouched regions randomly selected as distractors.
We use human error in grounding the edited regions as a proxy for the realistic and subtle nature of localized feature/attribute alterations achieved in our dataset.

\subsection{Crowdsourcing and Setup}

We hosted the two evaluation tasks as separate web apps and crowdsourced them through Cloud Research. 
For the first task, we prepared 400 total image triplets, each including two real images and one edited image.
Fake images for 200 of these triplets were randomly selected from our \emph{DETER}, and another 200 equally sampled from existing deep fake sources including SeqDeepFake~\cite{shao2022detecting}, DGM$^4$~\cite{shao2023detecting}, OpenForensics~\cite{le2021openforensics}, and DDPMs~\cite{ho2020dpm}.
For the second task, we had 100 triplets assembled using 100 photos from our dataset with random facial features/regions altered. 

\begin{figure*}[htb] 
    \centering
    \begin{minipage}{0.47\textwidth}
        \centering
        \includegraphics[width=\linewidth]{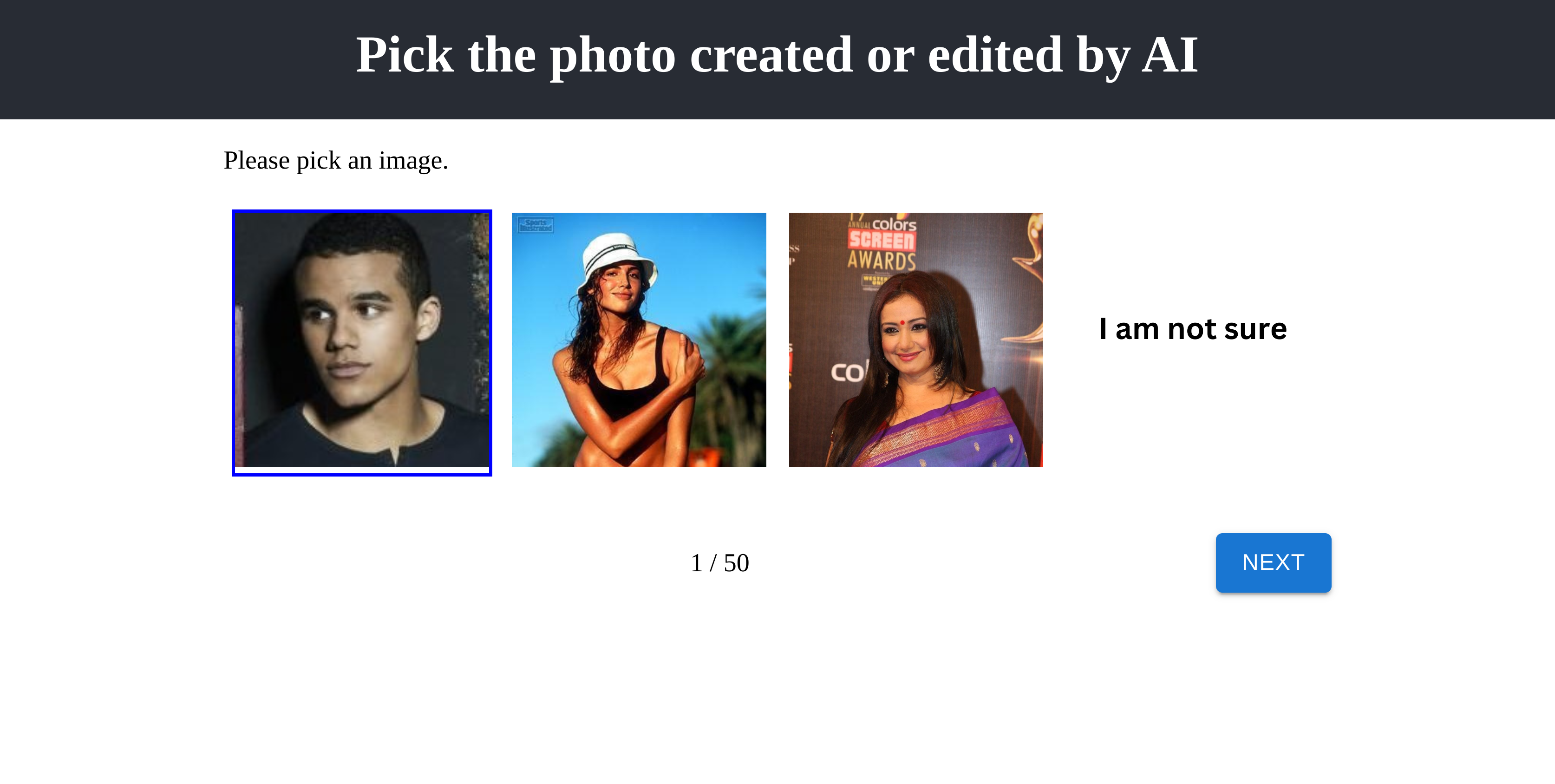}
        \caption{\textbf{Task layout for human evaluation - General Quality Assessment}}
        \label{fig:humanstudy-quality}
    \end{minipage}\hfill
    \begin{minipage}{0.47\textwidth}
        \centering
        \includegraphics[width=\linewidth]{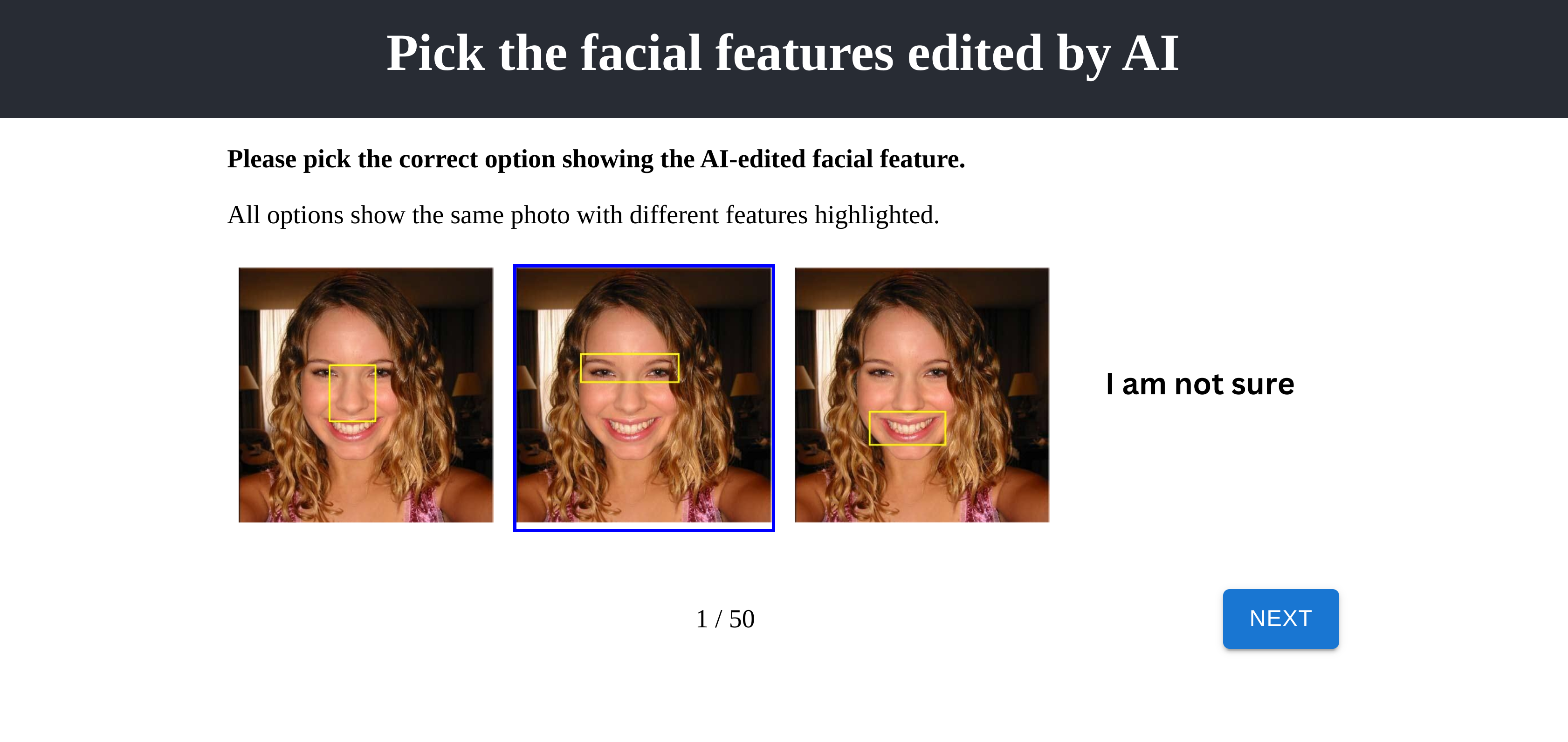}
        \caption{\textbf{Task layout for human evaluation - Regional Fake Detection}}
        \label{fig:humanstudy-region}
    \end{minipage}
\end{figure*}

For both tasks, in addition to three image options, we also included a \emph{``I am not sure''} option, which allows the evaluators to forfeit instead of forcing them to make a choice when it comes to hard samples.
The layout of the survey for one selection is shown in Figures \ref{fig:humanstudy-quality} and \ref{fig:humanstudy-region} respectively for tasks 1 and 2. 

\begin{figure*}[htb] 
    \centering
    \begin{minipage}{0.47\textwidth}
        \centering
        \includegraphics[width=\textwidth]{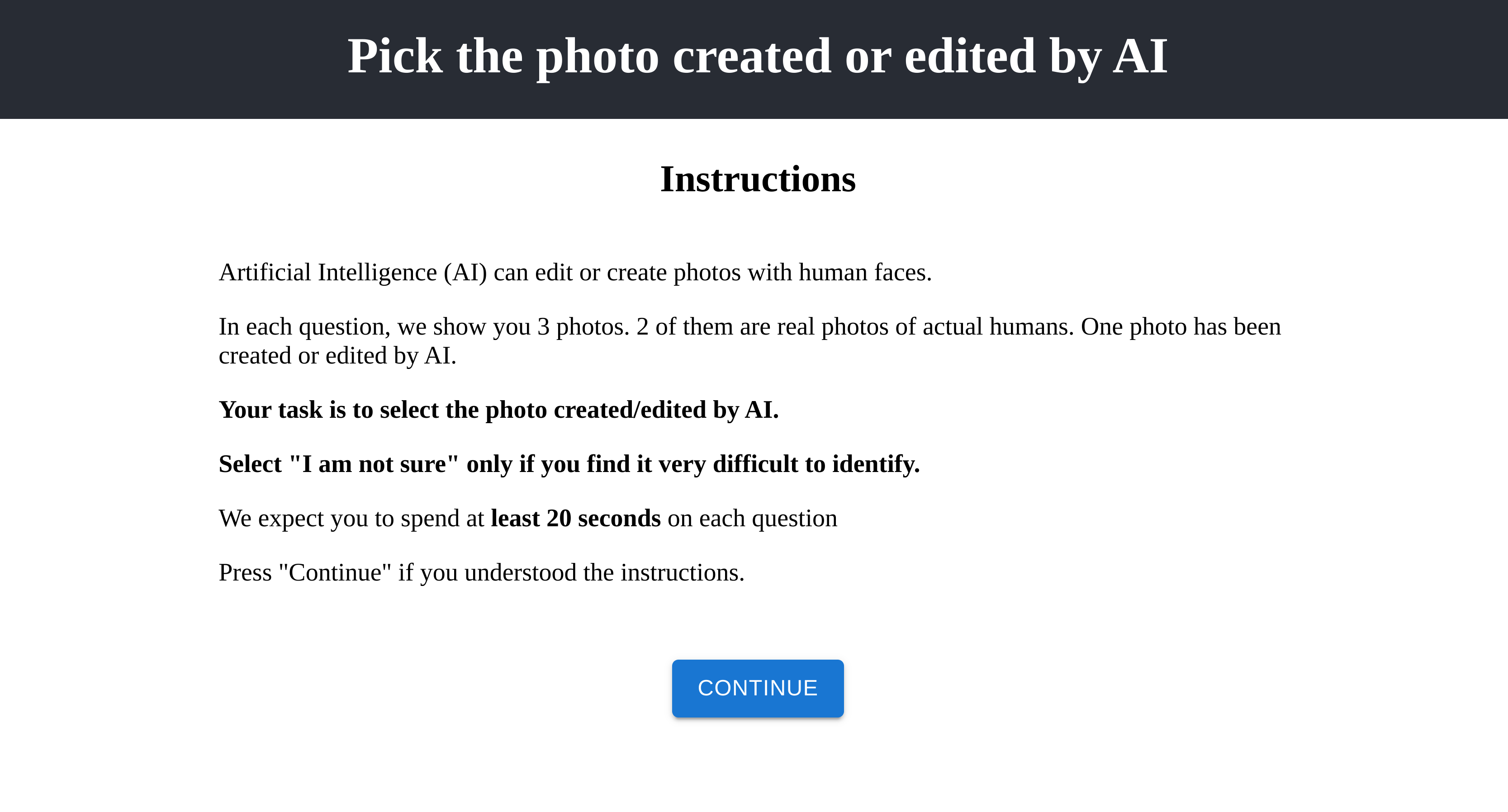}
        \caption{\textbf{Task instructions for human evaluation - General Quality Assessment}}
        \label{fig:humanstudy-quality-ins}
    \end{minipage}\hfill
    \begin{minipage}{0.47\textwidth}
        \centering
        \includegraphics[width=\textwidth]{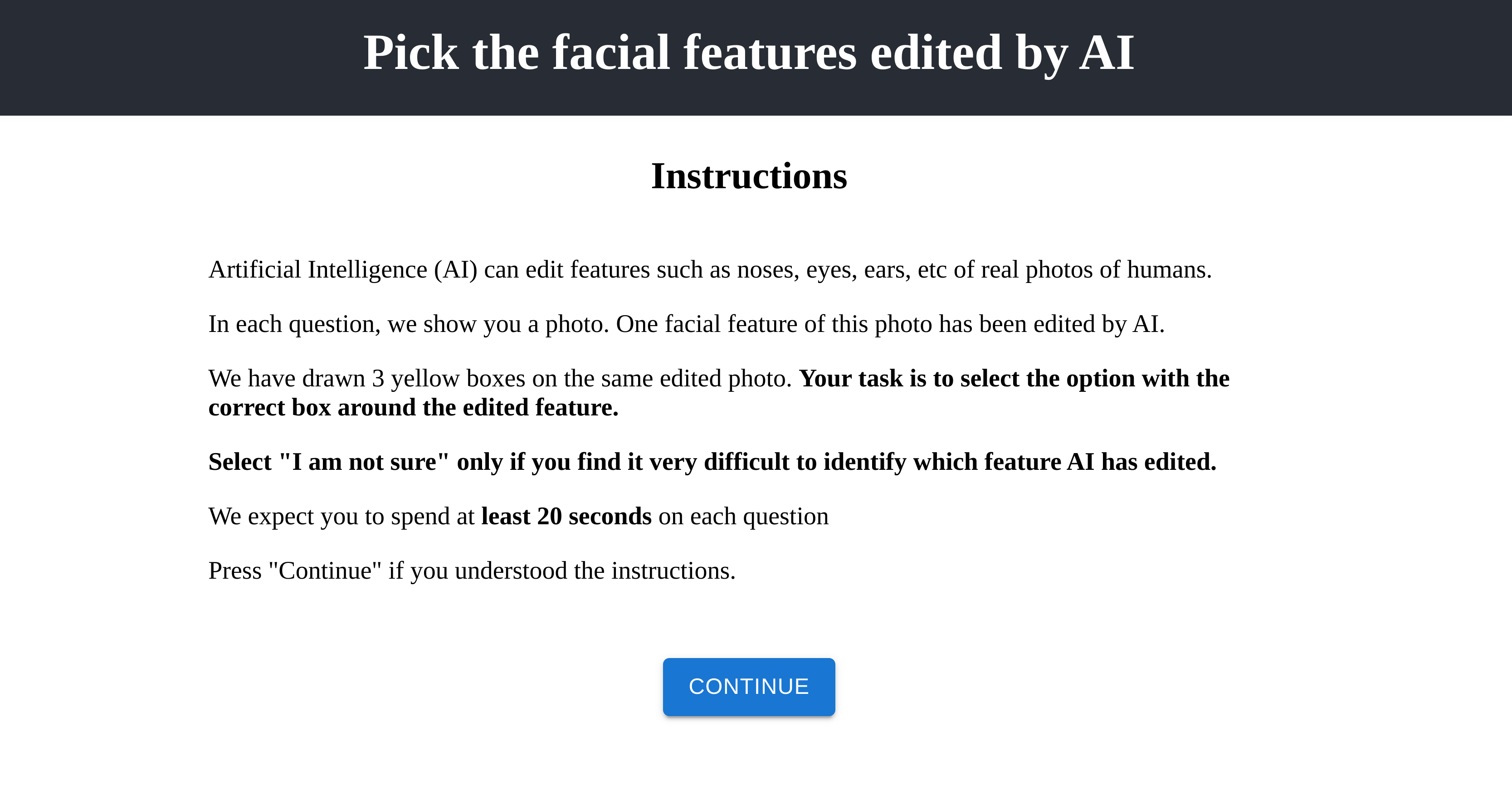}
        \caption{\textbf{Task instructions for human evaluation - Regional Fake Detection}}
        \label{fig:humanstudy-region-ins}
    \end{minipage}
\end{figure*}

For both tasks, we split our triples into multiple surveys containing 50 image triplets each. Each survey with  50 image triplets was completed by 3 human evaluators. To ensure that crowdsourced human evaluators spend adequate time looking for cues of deepfakes in each selection, we encourage them to spend at least 20 seconds on each selection. 
The instructions given to the evaluators for the two tasks are shown in Figure.\ref{fig:humanstudy-quality-ins} and \ref{fig:humanstudy-region-ins}.

\section{More Details about Regional Fake Detection}
\label{app_sec:detection}

We report the experimental results measured with IoU=0.5 in the main paper, and provide additional results with IoU=0.75 in Tab.~\ref{tab:appendix_results1}.
The additional results further validate and support our break-down analysis and take-away messages presented in the main paper.

\begin{figure*}[th]
    \centering
    \includegraphics[width=0.98\textwidth]{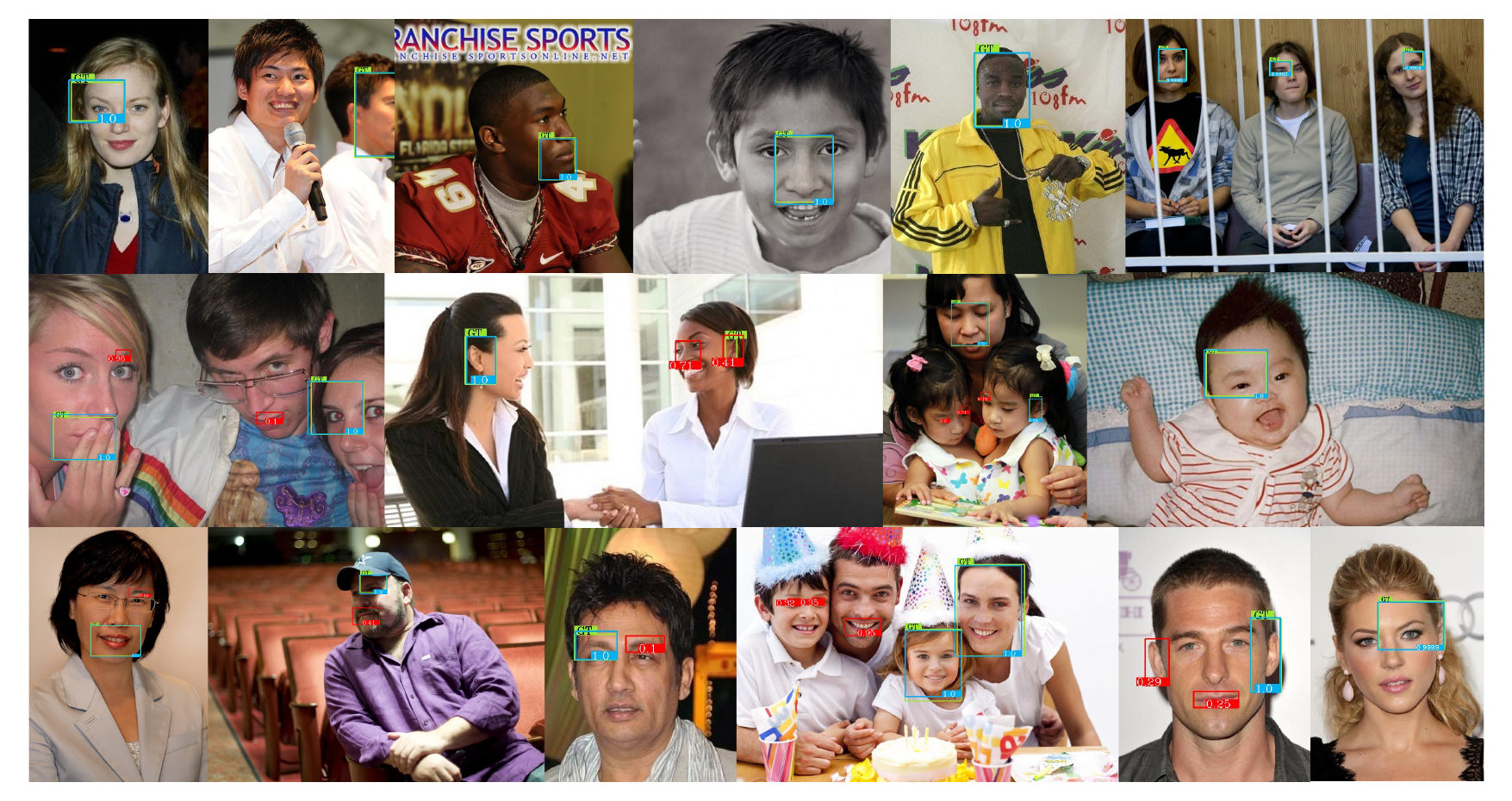}
    \caption{\textbf{Additional qualitative results of regional fake detection.} GT, correct predictions, and false positives are annotated in green, blue, and red boxes, respectively. Current models induce a relatively high false alarm rate.}
    \label{fig:appendix_qualitative}
\end{figure*}

As shown in Tab.~\ref{tab:appendix_results1}, the performance of various models uniformly decreases with the increasing stringency of IoU constraints, aligning with the overall conclusion of the main paper. Specifically, among the three tasks, inpainting exhibits the poorest performance. This is primarily attributed to the arbitrary shape of masks, in contrast to the relatively fixed mask transformation ranges in the other tasks, further underscoring the issue of spurious correlations during the dataset construction stage. In attribute editing, the modified regions are more fixed compared to face swapping, focusing on specific facial areas such as the eyes, mouth, and nose. Consequently, attribute editing achieves the highest precision. Despite its elevated recall, the precision across all tasks remains at a relatively low level. This discrepancy indicates that the model has biases in the learning process, where it fails to adequately capture the inherent differences between features in real and fake images, leading to a significant number of false positives. To address this issue, we introduce additional real images, i.e., improved settings in Tab.~\ref{tab:appendix_results1}, during the training process to encourage the model to better discern between real and fake images. This strategy results in a substantial improvement of over 20\% in precision and accuracy across all tasks and methods. Therefore, ensuring comprehensive learning of distinctions in features between real and fake images is a crucial focal point for advancing the task of fake regional detection.
Fig.~\ref{fig:appendix_qualitative} includes more qualitative samples.

We have also provided additional generator-based cross-domain results for both inpainting and attribute editing tasks. From Tab.~\ref{tab:appendix_gan_diff}, it is evident that the cross-domain performance of models trained with the GANs-based generators significantly surpasses those trained with the DMs-based generators, even outperforming the original DMs domain in inpainting tasks. Specifically, models trained with GANs-based generators exhibit superior performance on GANs-based and DMs-based test data (GANs + DMs), once again highlighting the robustness of features generated by GANs over DMs-based features. Additionally, there is complementary information in the features of GANs-based and DMs-based generators, and joint training further enriches the representation of fake features, leading to better results.

\begin{table*}[t]
\centering
\scalebox{0.65}{
\begin{tabular}{lcccccccccccccccc}
\toprule
Methods                     & \multicolumn{1}{c|}{}           & \multicolumn{3}{c|}{\begin{tabular}[c]{@{}c@{}}Classification\\ (image-level)\end{tabular}} & \multicolumn{9}{c|}{\begin{tabular}[c]{@{}c@{}}Object Detection\\ (box-level)\end{tabular}}                                                                                                                                 & \multicolumn{3}{c}{\begin{tabular}[c]{@{}c@{}}Instance Segmentation\\ (mask-level)\end{tabular}} \\ \hline
                            & \multicolumn{1}{c|}{Operations} & \multicolumn{1}{c|}{Swap}  & \multicolumn{1}{c|}{Inpaint}  & \multicolumn{1}{c|}{Attribute} & \multicolumn{3}{c|}{Swap}                                               & \multicolumn{3}{c|}{Inpaint}                                            & \multicolumn{3}{c|}{Attribute}                                          & \multicolumn{1}{c|}{Swap}           & \multicolumn{1}{c|}{Inpaint}          & Attribute          \\ \cline{2-17} 
                            & \multicolumn{1}{c|}{Setup}    & \multicolumn{3}{c|}{Accuracy}                                                               & Precision             & Recall                & \multicolumn{1}{c|}{AP} & Precision             & Recall                & \multicolumn{1}{c|}{AP} & Precision             & Recall                & \multicolumn{1}{c|}{AP} & \multicolumn{3}{c}{Mask AP}                                                                      \\ \hline \hline

\multirow{2}{*}{MaskR-CNN 17'~\cite{he2017mask}} & C.   & 0.51         & 0.40            & 0.40             & 0.24      &  \textbf{0.97}  &  \textbf{0.96} & 0.20      & 0.77   & 0.73 & 0.33      & 0.88   & 0.81 & 0.958           & 0.718              & 0.807               \\  
  & I.    & 0.75      &   0.65          &  0.62            & 0.45      &  0.96  &  0.95 &  0.35     & 0.77   & 0.74 &  0.49     & 0.86   & 0.82 &  0.954          &    \textbf{0.738}         &   \textbf{0.818}            \\  
  \hline
   \multirow{2}{*}{Yolact 19'~\cite{bolya2019yolact} }                &     C.        &  0.52            & 0.41               &   0.42      &  \underline{0.08}           & 0.96       & \textbf{0.96}     &  \underline{0.05}         & \underline{0.67}       & 0.60     &  \underline{0.08}        & 0.78       &  \underline{0.68}    & 0.955               & \underline{0.564}                  &   \underline{0.655}
   \\ 
& I.    &  0.85        &  0.73           &    0.71          &  0.46     & 0.96 & \textbf{0.96} & 0.27      & 0.69   & 0.64 & 0.39      &  \underline{0.76}  &  0.70 & \textbf{0.959}           &   0.617           &   0.685             \\  
     \hline
     \multirow{2}{*}{Mask2Former 22'~\cite{cheng2022masked} }                &  { C.}        &  0.47           & 0.38                 &  \underline{0.38}             & 0.20         &  0.96     &  \underline{0.94}   &  0.16      & 0.70       & \underline{0.56}    & 0.28         &  0.85      &  0.76    &  
    \underline{0.946}            & 0.578              &   0.758
           \\ 
       & I.    & 0.78         &   0.65          &     0.63      & 0.44      & 0.96   & 0.95 & 0.28      &  0.67  &0.61  &  0.44     & 0.82   & 0.75 & 0.953     & 0.638             &   0.735             \\  
\hline
                      \multirow{2}{*}{FasterR-CNN 15'~\cite{girshick2015fast} }                &      C.        &  0.53    & 0.40     & 0.39   & 0.27  &   \textbf{0.97 }& \textbf{0.96}  &   0.20    &  0.72   &  0.67 &   0.33     &  0.84      & 0.78    &         -      &    -              &   -
           \\ 
       & I.    &    0.77      &  0.66           & 0.63     &  0.50     & 0.96   & 0.95 &  0.35     &  0.73  & 0.69 &  0.50     & 0.83   & 0.78 &   -         & -             &   -             \\  
 \hline
                           \multirow{2}{*}{YOLOX 21'~\cite{ge2021yolox} }                &      { C.}        &  0.54           &  0.48               &   0.49            &   0.29        &  0.96      &  0.95   &    0.26     &  0.77        & 0.69      &    0.40      &  0.86      &   0.80   &  -             &   -               &   -
           \\ 
       &  I.    &     \textbf{0.92}  & \textbf{0.82}      &  \textbf{0.79}      &   \textbf{0.77}  & \underline{0.95}  & 0.95 & \textbf{0.58}      & 0.77   & 0.74 & \textbf{0.69}      & 0.81   & 0.79 &    -        &       -       &         -       \\  
\hline
                           \multirow{2}{*}{DINO 22'~\cite{zhang2022dino} }                &      {C.}        &         \underline{0.44}    & \underline{0.36}            &  0.40  &        0.11    &  \textbf{0.97}           &   \textbf{0.96}   & 0.09    & 0.78       &   0.72     &   0.18   & \textbf{0.90}         &     0.82   &    -                &   -               &  - 
           \\ 
       & I.    & 0.74         &   0.65          &  0.65            &   0.28    & \textbf{0.97}    &  \textbf{0.96} & 0.19      & \textbf{0.79}   & \textbf{0.75} &   0.33    &  \textbf{0.90}  & \textbf{0.85} &        -    &          -    &       -         \\  
\bottomrule
\end{tabular}}
\caption{\textbf{Quantitative evaluation results for regional fake detection under \emph{(C)}onventional (i.e., training w/o negative image samples) and \emph{(I)}mporved (i.e., training with negative image samples) settings  with IoU=0.75.} 
All metrics are the higher the better, best and worst results are marked in \textbf{bold} and \underline{underlined}, respectively.}
\vspace{-0.1in}
\label{tab:appendix_results1}
\end{table*}

\begin{table*}[t]
\centering
\scalebox{0.65}{
\begin{tabular}{c|cccccc|cccccc|cccccc}
\hline
           & \multicolumn{6}{c|}{GANs}                                                       & \multicolumn{6}{c|}{DMs}                                                        & \multicolumn{6}{c}{GANs + DMs}                                                 \\ \cline{2-19} 
           & \multicolumn{3}{c|}{Inpaint}                   & \multicolumn{3}{c|}{Attribute} & \multicolumn{3}{c|}{Inpaint}                   & \multicolumn{3}{c|}{Attribute} & \multicolumn{3}{c|}{Inpaint}                   & \multicolumn{3}{c}{Attribute} \\
           & Precision & Recall & \multicolumn{1}{c|}{AP}   & Precision   & Recall   & AP    & Precision & Recall & \multicolumn{1}{c|}{AP}   & Precision   & Recall   & AP    & Precision & Recall & \multicolumn{1}{c|}{AP}   & Precision   & Recall  & AP    \\ \hline
GANs       & 0.48      & 0.9    & \multicolumn{1}{c|}{0.87} & 0.56        & 0.94     & 0.91  & 0.48      & 0.91   & \multicolumn{1}{c|}{0.88} & 0.42        & 0.79     & 0.67  & 0.48      & 0.91   & \multicolumn{1}{c|}{0.87} & 0.50        & 0.88    & 0.82  \\
DMs        & 0.38      & 0.76   & \multicolumn{1}{c|}{0.71} & 0.43        & 0.78     & 0.64  & 0.53      & 0.92   & \multicolumn{1}{c|}{0.89} & 0.59        & 0.95     & 0.92  & 0.44      & 0.83   & \multicolumn{1}{c|}{0.79} & 0.50        & 0.85    & 0.77  \\
GANs + DMs & 0.52      & 0.91   & \multicolumn{1}{c|}{0.88} & 0.58        & 0.95     & 0.91  & 0.55      & 0.93   & \multicolumn{1}{c|}{0.91} & 0.59        & 0.95     & 0.92  & 0.53      & 0.92   & \multicolumn{1}{c|}{0.89} & 0.58        & 0.95    & 0.91  \\ \hline
\end{tabular}
}
\caption{\textbf{Quantitative results in terms of GANs-based and DMs-based generators in cross-domain experiments with Mask R-CNN~\cite{he2017mask}.} The scores are calculated with IoU=0.5.}
\label{tab:appendix_gan_diff}
\end{table*}




\end{document}